\documentclass{article}

\usepackage{natbib}
\usepackage{colortbl}

\usepackage[T1]{fontenc}
\usepackage{graphicx}
\usepackage[misc]{ifsym}

\usepackage[table]{xcolor} 
\usepackage[T1]{fontenc}
\usepackage{graphicx}

\usepackage{amsfonts}
\usepackage{amsmath}
\usepackage{pgfplots}
\pgfplotsset{compat=1.17}
\usepackage{bm}
\usepackage{amssymb}
\usepackage{graphicx}
\usepackage{tikz}
\usetikzlibrary{shapes}
\usepackage{subcaption}
\usepackage[utf8]{inputenc}
\usepackage{multirow}
\usepackage{array, makecell, rotating}
\usepackage{blindtext}

\usepackage{listofitems} 
\usepackage{booktabs}
\usepackage[table]{xcolor}
\usepackage{adjustbox}

\usepackage{color}
\usepackage{amsmath}
\usepackage{hyperref}

\newcommand{\vect}[1]{\mathbf{#1}}

\newcommand{\kl}[2]{D_{KL} \left( #1 \mid \mid #2 \right)}

\newcommand{\expected}[1]{\mathop{\mathbb{E}}_{\substack{#1}}}

\newcommand{\naturalset}[0]{\mathbb{N}}

\newcommand{\sample}[2]{#1 \sim #2}

\newcommand{\latentspaceconstraintclosedformNoI}[0]{ \frac{1}{2} \sum_{i=1}^{D} \left( -\log \sigma_i^2 - 1 + \sigma_i^2 +  \mu_i^2 \right) }

\newcommand{\mufat}[0]{\bm{\mu}}
\newcommand{\sigmafat}[0]{\bm{\sigma}}

\newcommand{\epilonfat}[0]{\bm{\epsilon}}

\newcommand{\identitymtx}[0]{\mathbf{I}}

\newcommand{\diagsigma}[0]{\text{diag}(\sigmafat)} 

\newcommand{\standardnormal}[0]{\mathcal{N}(\bm{\vect{0}}, \identitymtx)}

\newcommand{\tildex}[0]{\widetilde{\mathbf{x}}}

\newcommand{\samplestandardnormal}[1]{\sample{#1}{\standardnormal}}

\newcommand{\x}[0]{\vect{x}}
\newcommand{\xt}[0]{\x_t}

\newcommand{\z}[0]{\vect{z}}
\newcommand{\zt}[0]{\z_t}

\newcommand{\zj}[0]{\z_j}
\newcommand{\ztk}[0]{\vect{z}_{t+k}}
\newcommand{\ztone}[0]{\vect{z}_{t+1}}

\newcommand{\zmstart}[0]{\z^m_{\text{start}}}
\newcommand{\zmtarget}[0]{\z^m_{\text{target}}}

\newcommand{\zbidi}[0]{\z^3_{\text{bidi}}}
\newcommand{\zbaga}[0]{\z^3_{\text{baga}}}

\newcommand{\zmalpha}[0]{\z^m_{\alpha = 1}}

\newcommand{\ct}[0]{\vect{c}_t}

\newcommand{\fkmblank}[0]{f_k^m(\cdot)}

\newcommand{\fkm}[0]{f_k^m({\ztk^m},{\zt^m})}
\newcommand{\fkmMonteCarlo}[0]{f_k^m({\ztk^m}^{(l)},{\zt^m}^{(l)})}

\newcommand{\gim}[0]{\log \frac{\fkm }{\sum_{\zj^m \in X} f_k^m(\zj^m, \zt^m)}}

\newcommand{\gimMonteCarlo}[0]{\log \frac{\fkmMonteCarlo}{\sum_{\zj^m \in X} f_k^m(\zj^m, {\zt^m}^{(l)})}}

\newcommand{\qzzblank}[0]{q^m(\cdot \mid \zt^{m-1})}

\newcommand{\latentspaceconstraintgim}[0]{\kl{ \qzzblank }{ \standardnormal } }

\newcommand{\sampleqmdot}[1]{q^m(\cdot \mid #1)}

\newcommand{\reconstrgim}[0]{-\sum_k
\expected{
	\sample{\ztk^m}{\sampleqmdot{\ztk^{m-1}} } \\ 
	\sample{\zt^m}{\sampleqmdot{\zt^{m-1}}} } 
\left[ \gim \right]
}

\newcommand{\reconstrgimMontecarlo}[0]{-\sum_k
	\frac{1}{L} 
	\left[\sum_{l=1}^L \gimMonteCarlo \right]
}

\newcommand{\gencm}[0]{g_{enc}^m(\cdot)}

\newcommand{\gar}[0]{g_{ar}(\cdot)}

\newcommand{\Lnce}[0]{\mathcal{L}_{\text{NCE}}}
\newcommand{\Lsim}[0]{\mathcal{L}_{\text{S-NCE}}}

\newcommand{\qfromzmneg}[0]{\sampleqmdot{\zt^{m-1}}}

\newcommand{\normalfatmusigma}[0]{\mathcal{N}(\mufat, \diagsigma)}

\newcommand{\Z}[0]{\mathcal{Z}}

\newcommand{\ztm}[0]{\zt^{m}}
\newcommand{\ztmnegone}[0]{\zt^{m-1}}

\usepackage{PRIMEarxiv}
\usepackage[utf8]{inputenc} 
\usepackage[T1]{fontenc}

\usepackage{hyperref}       
\usepackage{url}            
\usepackage{booktabs}       
\usepackage{amsfonts}       
\usepackage{nicefrac}       
\usepackage{microtype}      
\usepackage{lipsum}
\usepackage{fancyhdr}       
\usepackage{graphicx}       
\graphicspath{{media/}}     

\title{\large{Smooth InfoMax – Towards Easier Post-Hoc Interpretability}%
\thanks{This paper was accepted for publication at the European Conference on Machine Learning and Principles and Practice of Knowledge Discovery in Databases (ECML-PKDD 2025), held on September 15–19, 2025 in Porto, Portugal.}}

\author{
    Fabian Denoodt\\
    University of Antwerp, IDLab-imec, sqIRL\\
    \texttt{fabian.denoodt@uantwerpen.be}\\
    \And
    Bart de Boer\\
    Vrije Universiteit Brussel\\
    Artificial Intelligence Lab\\
    \texttt{bart@ai.vub.ac.be}\\
    \And
    José Oramas\\
    University of Antwerp, IDLab-imec, sqIRL\\
    \texttt{jose.oramas@uantwerpen.be}\\
}

\begin{document}
    \maketitle
    \begin{abstract}
We introduce Smooth InfoMax (SIM), a self-supervised representation learning method that incorporates interpretability constraints into the latent representations at different depths of the network. Based on $\beta$-VAEs, SIM's architecture consists of probabilistic modules optimized locally with the InfoNCE loss to produce Gaussian-distributed representations regularized toward the standard normal distribution. This creates smooth, well-defined, and better-disentangled latent spaces, enabling easier post-hoc analysis. Evaluated on speech data, SIM preserves the large-scale training benefits of Greedy InfoMax while improving the effectiveness of post-hoc interpretability methods across layers.
    
Our code is available via \href{https://github.com/fdenoodt/Smooth-InfoMax}{\underline{GitHub}}.

\end{abstract}

    \section{Introduction}
Black-box models, particularly deep neural networks (NNs), have shown remarkable performance in recent years. However, despite their impressive success, their lack of interpretability poses a significant challenge, limiting their use in high-stakes decision environments. Consequently, various post-hoc interpretability techniques have been explored. Notable contributions include the work of~\cite{simonyanDeepConvolutionalNetworks2014}, which aims to find the input image that maximally activates a specific neuron in the network, and the research by~\cite{zeilerVisualizingUnderstandingConvolutional2013}, which focuses on highlighting the regions in the input that a particular neuron is sensitive to.

However, the effectiveness of these post-hoc methods decreases in complex models due to the large number of neurons that must be analyzed. Additionally, as argued by~\cite{bauNetworkDissectionQuantifying2017}, the internal semantic concepts learned by these neurons are typically highly entangled throughout the network. This makes the interpretation of a neuron particularly difficult, as multiple neurons may work as a whole and together be sensitive to a given semantic concept while other neurons may not be contributing anything at all. For these reasons, it is likely impossible to fully understand these NNs with just the existing post-hoc interpretability techniques. In contrast, inherently interpretable models (e.g., logistic regression and decision trees) offer more transparency but may struggle with complex problems.

Another challenge with NNs is that they are typically trained end-to-end, which requires significant memory, especially as models grow larger. This can pose hardware constraints, as training must fit within the available memory of the device. Additionally, deeper networks can be more susceptible to the vanishing gradient problem~\cite{hochreiter_vanish_gradient}.

To address these issues, we propose Smooth-InfoMax, a self-supervised representation learning method that integrates two existing paradigms: Greedy InfoMax (GIM)~\cite{lowePuttingEndEndtoEnd2020a} and $\beta$-Variational Autoencoders ($\beta$-VAEs)~\cite{burgessUnderstandingDisentanglingBeta2018}. This integration improves the post-hoc interpretability of the NN while enabling large-scale distributed training, combining benefits not achievable by either paradigm alone.

SIM's learning objective is based on contrastive learning and does not require labels or a decoder for training. Building upon GIM, SIM splits the architecture into modules, each trained greedily with a novel loss based on the InfoNCE bound~\cite{oordRepresentationLearningContrastive2019}. As such, we preserve benefits such as large-scale distributed training of architectures that would otherwise not fit in memory and reduced vanishing gradients issues~\cite{lowePuttingEndEndtoEnd2020a}. 

Furthermore, SIM incorporates the latent-space regularization properties of $\beta$-VAEs across various depths in the network. This helps create smooth and well-structured latent spaces that encourage disentanglement~\cite{burgessUnderstandingDisentanglingBeta2018, sikkaCloserLookDisentangling2019, higginsBetaVAELearningBasic2022}. As a result, small changes in the latent space correspond to small changes in the input space, making post-hoc interpretability easier. However, unlike $\beta$-VAEs, SIM does not require a decoder during training, reducing memory usage. Another key difference is that SIM applies this regularization across different layers, making it easier to analyze representations throughout the network, rather than $\beta$-VAEs where the regularization typically is only applied at a single layer. A decoder can then be used as a post-hoc interpretability tool by traversing a latent space in the network, revealing the information that a particular neuron is sensitive to.  Obtaining meaningful insights with such a procedure would be a lot harder if the spaces were not as well structured, as is typically the case in conventional NNs~\cite{doersch2016tutorial, bengio2013representation}.

Our contributions are the following:
\begin{enumerate}
    \item Introducing SIM, a framework with a novel loss function and probabilistic architecture for easier interpretable latent spaces, evaluated on sequential speech data. Although a relatively straightforward integration of existing methods, this proposed combination provides specific benefits not achievable by either approach alone.
    \item  We show, via a decoder, that SIM produces latent spaces that are easier to analyze. This also leads to a new metric for quantifying the number of dimensions required for successful reconstructions. 
    \item Empirically showing that ideas from $\beta$-VAE extend to other frameworks and can be repeated at different depths without significant performance loss.
\end{enumerate}
\textbf{Reproducibility:} Our code and commands to replicate the experiments are all available via \href{https://github.com/anonymoususerforpeerreview/Smooth-InfoMax}{\underline{GitHub}}.

    \section{The Starting Point - Greedy InfoMax}
Greedy InfoMax (GIM) learns representations from sequential data without the need for labels by exploiting the assumption of slowly varying data~\cite{wiskott_slow_features}. This assumption is for instance applicable to speech signals where the conveyed information at time step $t$ and $t+k$ contains redundancy, such as the speaker's identity, the conveyed emotion and the pronounced phonemes~\cite{lowePuttingEndEndtoEnd2020a}. Meanwhile, this information may not necessarily be shared with random other patches of speech.
An encoder can then be optimized to create representations that maximally preserve the shared information between the representations of temporally nearby patches~\cite{lowePuttingEndEndtoEnd2020a}, while at the same time discarding low-level information and noise that is more local~\cite{oordRepresentationLearningContrastive2019}. It has been shown that such a strategy creates highly competitive representations for downstream tasks in various domains~\cite{henaffDataEfficientImageRecognition2020, oordRepresentationLearningContrastive2019, lowePuttingEndEndtoEnd2020a, stackeEvaluationContrastivePredictive2020, NhemDWPOP25, luSemiSupervisedHistologyClassification2019, bhatiSegmentalContrastivePredictive2021}.

\subsubsection{The network architecture}
An audio sequence is split up into patches $\x_1 \dots \x_T$ where each $\xt$ is a vector of fixed length, containing for instance 10ms of speech. Each patch $\xt$ is encoded by passing it through a series of $M$ encoder modules: $g_{enc}^1(\cdot),~ g_{enc}^2(\cdot),~\dots,~g_{enc}^M(\cdot)$. An encoder module consists of one or more convolution layers. The final representation $\zt^M$ is then obtained by propagating $\xt$ through each module as follows: 
\begin{equation}
    g_{enc}^M ( \dots	g_{enc}^2(g_{enc}^1(\xt))) = \zt^M .
\end{equation} 
As such, each module's output is the input of the successive module: $g_{enc}^m(\zt^{m-1}) = \zt^m$. For tasks where context-related information is required, the final module $g_{enc}^M$ can be replaced by an autoregressive module $g_{ar}(\z_1^{M-1} ~ \dots ~ \zt^{M-1}) = \ct$. The autoregressive module can for instance be represented as a Gated Recurrent Unit (GRU). Both $\zt^M$ or $\ct$ may serve as the representation for downstream tasks and can be pooled into a single vector if needed.

\subsubsection{The loss function}
Given a representation $\ztm$ and a set $X =  \left\{ \z_1^m, \z_2^m, \dots \right\}~\cup~$ $\left\{ \ztone^m, \dots, \z_{t+K}^m \right\}$ consisting of random encoded audio patches and $K$ subsequent samples of $\ztm$, respectively, GIM learns to preserve the information between temporally nearby representations by learning to discriminate the subsequent \textit{positive} samples $\ztk^m$ from the \textit{negative} random samples $\zj^m$ using a function $f_k^m(\cdot)$ which scores the similarity between two latent representations~\cite{lowePuttingEndEndtoEnd2020a}. This function is defined as follows:
\begin{align} 
	f_k^m(\ztk^m,\zt^m) = \exp({\ztk^m}^TW_k^m\zt^m), \label{eq:fkm}
\end{align}
where $W_k$ is a weight matrix which is learned. Intuitively, due to the slowly varying data assumption, the similarity score for positive patches should be high and small for negative patches. The InfoNCE loss, used to optimize an \textit{individual} module $\gencm$ and its respective $W_k^m$ is shown below:
\begin{equation} 
	\Lnce^m = -\sum_{k} \expected{\textsubscript{X}} \left[\log \frac{\fkm }{\sum_{\zj^m \in X} f_k^m(\zj^m, \zt^m)} \right].	
\end{equation}
One can prove that minimizing the InfoNCE loss is equivalent to maximizing a lower bound on the mutual information between $\ztm$ and $\ztk^m$~\cite{oordRepresentationLearningContrastive2019}:
\begin{equation}
	I(\ztk^m; \zt^m) \ge \log(N) - \Lnce^m.
\end{equation}

As a result of GIM's greedy approach, a conventional neural network architecture can be divided into modules. These modules can be trained either in parallel on distributed devices or sequentially, enabling the training of models larger than device memory and reducing the vanishing gradient problem. In the following section, we discuss how we can preserve these benefits in SIM, while also allowing for better interpretability.

    \section{Smooth InfoMax}
While optimizing for the InfoNCE bound, as done in GIM, is remarkably successful for downstream classification, analyzing the learned representations remains difficult. In what follows we introduce Smooth InfoMax (SIM), maintaining the computational benefits obtained from optimizing the InfoNCE objective, while introducing easily traversable latent spaces and better disentangled representations at different depths in the network due to techniques borrowed from $\beta$-VAEs.

\subsection{Towards Decoupled Training for Probabilistic Representations} \label{cha:vgim_decoupled_training_for_probabil_repr}
The architecture is again based on modules, where the modules $g_{enc}^1(\cdot),~ g_{enc}^2(\cdot),~$ $\dots,~g_{enc}^M(\cdot)$ are each greedily optimized without gradients flowing between them. However, rather than producing a single deterministic point $\ztm$, the output from $\gencm$ is now a multivariate Gaussian distribution $q(\zt^m \mid \zt^{m-1})$, parameterized by the mean vector $\mufat$ and covariance matrix $\diagsigma$. More precisely, we have:
\begin{equation}
    g_{enc}^m(\ztmnegone) = q(\zt^m \mid \zt^{m-1}) = \normalfatmusigma,
\end{equation}
with $\mufat$ and $\sigmafat$ dependent on $\zt^{m-1}$. A point $\zt^m$ is then obtained by sampling from this distribution, denoted respectively, as follows:
\begin{align} 
    \sample{\zt^m}~ & \qfromzmneg  \label{eq:sample_z_from_q}.
\end{align}
The encoding modules are thus stochastic and obtaining two representations from the same input will not necessarily produce the same result. This is in contrast to GIM's latent representations which remain fixed with respect to the input.

We obtain these stochastic modules by defining each module $\gencm$ consisting of two blocks. The first block receives as input $\zt^{m-1}$ and predicts the parameters $\mufat$ and $\sigmafat$. The second block samples $\sample{\zt^m} \sampleqmdot{\zt^{m-1}}$ from this distribution and produces an output representation. In practice, sampling from $q^m$ is achieved through a reparameterization trick, as introduced in~\cite{kingmaAutoEncodingVariationalBayes2022}. The equation to compute $\zt^m$ then becomes:
\begin{equation*}
    \zt^m = \mufat + \sigmafat \odot \epilonfat,
\end{equation*}
where $\epilonfat$ corresponds to a sampled value $\samplestandardnormal{\epilonfat}$ and $\odot$ is element-wise multiplication. The two blocks are depicted in Figure~\ref{fig:single_variational_module}. The optional autoregressive module $\gar$ has been untouched, and remains identical as in GIM, resulting in deterministic representations.
			
\begin{figure}
	\centering
	\tikzstyle{arrow} = [thick,->,>=stealth]
	\begin{tikzpicture}[scale=0.8,
		AnnNode/.style={trapezium, draw=black,
			trapezium stretches=true,
			minimum width=2cm, 
			minimum height=1.5cm,
			rotate=-90,
			trapezium angle=75,
			very thick},
		SamplingBlock/.style={rectangle, draw=black,
			minimum height=1cm, minimum width=2cm,
			very thick},
		]
		
		\node[AnnNode] (enc) {\rotatebox{90}{ANN}};
		\node[SamplingBlock] [right of=enc, xshift=2.5cm] (sample) {$\zt^m \sim \sampleqmdot{\zt^{m-1}}$};
		
		\draw[->] ++(-2.5, 0) -- (enc.south) node[above, midway] {$\zt^{m-1}$};
		\draw[->] 
		[transform canvas={yshift=.7em}] 
		(enc.north) -- (sample.west) node[above, midway] {$\mufat$};
		
		\draw[->] 
		[transform canvas={yshift=-.7em}] 
		(enc.north) --  (sample.west) node[below, midway] {$\sigmafat$};
		
		\draw[->] 
		(sample.east) --  ++(1.5, 0) node[above, midway] {$\zt^m$};

	\end{tikzpicture}
	\caption{A single module.}
	\label{fig:single_variational_module}
\end{figure}

\subsection{The Loss Function} \label{cha:vgim_learning_objective}
Instead of training the NN's modules end-to-end with a global loss function, each module is optimized greedily with its own loss. Through the introduction of the \textit{Smooth-InfoNCE} loss, mutual information between temporally nearby representations is maximized, while regularizing the latent space to be approximate to the standard Gaussian $\standardnormal$. This loss is defined as follows:
    {\tiny
\begin{align}
    \Lsim^m &= \underbrace{\reconstrgim}_{\text{Maximize } I(\ztk^m, \zt^m)} + \underbrace{\beta ~ \latentspaceconstraintgim}_{\text{Regularisation}}.
\end{align}
}
Here, $m \in \naturalset$ refers to the $m$'th module and $k \in \naturalset$ the number of follow-up patches the similarity score $\fkm$ must rate. The latent representations $\ztk^m$ and $\zt^m$ are encoded samples produced by $g_{enc}^m(\ztk^{m-1})$ and $g_{enc}^m(\zt^{m-1})$, respectively and $X$ is a set of samples ${ \left\{ \ztk^m, \z_1^m, \z_2^m, \dots \right\} }$ where $\zj^m$ with $j \neq t \! + \! k$ are random samples. In practice, the set can be based on the training batch. The parameter $\beta \ge 0$ is a hyper-parameter indicating the relative importance between the two terms. When $\beta = 0$, SIM is identical to GIM but with an altered architecture supporting probabilistic representations. The similarity score $f_k^m(\cdot)$ remains identical as in GIM:
\begin{equation}
    f_k^m(\ztk^m,\zt^m) = \exp({\ztk^m}^TW_k^m\zt^m).
\end{equation}

$\Lsim^m$ consists of two terms. The first term ensures that latent representations of temporally nearby patches maximally preserve their shared information. The second pushes the latent representations close to the origin.

\subsubsection{The Gradient}
To estimate the expectation term in $\Lsim$, we apply the same approximation method as in VAEs, achieved through Monte Carlo estimates~\cite{kingmaAutoEncodingVariationalBayes2022}. The first term in $\Lsim$ then becomes:
\begin{align*}
    \reconstrgimMontecarlo .
\end{align*}
Here, $L$ refers to the number of samples drawn. Each ${\ztk^m}^{(l)}$ and ${\zt^m}^{(l)}$ are different samples produced by their respective distributions. However, similar to~\cite{kingmaAutoEncodingVariationalBayes2022}, we can set $L=1$ without significantly hurting performance.

With regards to the second term in $\Lsim$, since $\qfromzmneg$ is a Gaussian defined by parameters $\mufat$ and $\sigmafat$, a closed-form solution exists~\cite{kingmaAutoEncodingVariationalBayes2022}. The closed-form equation is the following:
\begin{equation*}
    \latentspaceconstraintgim = \latentspaceconstraintclosedformNoI .
\end{equation*}
The variable $D$ refers to the number of dimensions of $\mufat$ and $\sigmafat$. This term can thus, be directly computed and does not need to be approximated through Monte Carlo estimates. The gradient for the two terms can then be computed using automatic differentiation tools such as PyTorch.

\subsection{Properties of the Latent Space} \label{cha:contin_space}
Here, we present two conjectures regarding the structure of the latent space defined by each of SIM's modules. They will serve as the main argument for why SIM's representations are more easily analyzable. Meanwhile, alternative contrastive approaches such as GIM lack these benefits. \\

\noindent \textbf{Conjecture 1.} \textit{$\Lsim$ enforces an uninterrupted and well-covered space around the origin.} \\
In SIM, a latent representation $\zt^m \in \Z^m$ of a data point $\zt^{m-1} \in \Z^{m-1}$ is a sample from a Gaussian distribution. Thus, encoding the same $\zt^{m-1}$ an infinite number of times results in a spherical region (around a particular mean $\mufat$) in $\Z^m$ that is covered by the latent representations corresponding to $\zt^{m-1}$, without any interruptions in this region. This is different from GIM where a data point merely covers a single point of the latent space (and not an entire region). Furthermore, because the KL divergence requires each region to be close to the origin, the regions are more likely to utilize the limited space efficiently around the origin, resulting in a lower chance of obtaining gaps between two regions from different data points.  \\

\noindent \textbf{Conjecture 2.} \textit{$\Lsim$ enforces smooth and consistent transitions in the latent space with respect to the shared information between temporarily nearby patches.} \\
The argument on why this holds true is similar to the argument made for VAEs~\cite{kingmaAutoEncodingVariationalBayes2022}. In the case of a VAE, a smooth space implies that a small change to $\z$ should result in a small change to its corresponding reconstruction, such that:
\begin{align}
    \z \approx \z' \implies p(\x \mid \z) \approx p(\x \mid \z').
\end{align}
Indeed, one can observe that the KL-divergence will encourage the region of latent points that a data point $\x$ can map to to be large. Meanwhile, the reconstruction error in a VAE encourages all the latent points falling in this region to be as close as possible to the initial data point $\x$. In SIM, the same argument can be used to obtain:
\begin{align}
    \ztm \approx {\ztm}' \implies f(\ztk^m, \ztm) \approx f(\ztk^m, {\ztm}'),
\end{align}
resulting in a smooth space with respect to the shared information between temporally nearby patches. Additionally, if a decoder is trained on SIM's representations, for the same reason, we obtain:
\begin{align}
    p(\xt \mid \ztm) \approx p(\xt \mid {\ztm}').
\end{align}

\subsubsection{Traversability of the space}
As a result of the smooth and well-defined shape, one can make small changes to $\ztm$ and observe what happens through a decoder with a much smaller risk of having abrupt changes to the corresponding $\x$, or obtaining out-of-distribution latent points that correspond to non-meaningful reconstructions due to gaps in the latent space. This results in an easily traversable latent space with a predictable structure, which is not guaranteed in conventional NNs, as they typically do not enforce these additional constraints.

\subsubsection{Disentanglement}
GIM poses no direct constraints on disentanglement risking having many dimensions of the representation together contribute a small amount to the contained information of an individual concept. However, as argued by~\cite{higginsBetaVAELearningBasic2022}, setting the prior $p(\vect{z})$ of the $\beta$-VAE's loss to an isotropic Gaussian encourages disentanglement in the representations. This results in each dimension from the encoding to capture a different property of the original data. In the case of $\Lsim$, the prior corresponds to the standard normal $\standardnormal$, and thus, this theorem is also applicable to SIM, and choosing a large value for $\beta$ in $\Lsim$ applies more pressure for the representations to be better disentangled.

    \section{Experiments and Evaluation}
We evaluate SIM's latent representations on raw speech data and compare them against GIM as a baseline. The goal is to measure the impact of $\beta$-VAE regularization in terms of raw performance and to analyze how effectively post-hoc interpretability techniques would perform. Since this work uses the well-established properties of $\beta$-VAEs, we do not aim to independently revalidate them. $\beta$-VAE's disentanglement, in particular, has been extensively confirmed~\cite{kim2018disentangling, higginsBetaVAELearningBasic2022, chen2018isolating} and benchmarked against other methods in prior work~\cite{locatello2019challenging}.

\subsection{Setup} \label{cha:experim_details_vgim}

\subsubsection{Dataset} We use two publicly available speech datasets. The first is an artificial dataset\footnote{Available at \url{https://github.com/fdenoodt/Artificial-Speech-Dataset}.} with a known and predictable structure, chosen because it provides a clear expectation of what a learning system should capture. The second, following GIM, is the 100-hour subset of the large-scale LibriSpeech dataset~\cite{lowePuttingEndEndtoEnd2020a, panayotov2015librispeech}.
The artificial dataset contains 851 fixed-length (640 ms) audio files, sampled at 16 kHz and split into 80\% training and 20\% test sets. Each file consists of a single spoken sound composed of alternating consonants and vowels (e.g., ``gi-ga-bu'').

\subsubsection{Architecture} SIM's architecture consists of three probabilistic CNN-based encoder modules and one autoregressive GRU module. Each CNN layer in these modules has 512 hidden dimensions. In $g_{enc}^1$, two convolutions (kernel: 10, 8; stride: 5, 4; padding: 2) are followed by parallel $\mu$ and $\sigma$ convolution layers (kernel: 1, stride: 1, no padding). $g_{enc}^2$ contains two convolutions (kernel: 4; stride: 2; padding: 2), followed by $\mu$ and $\sigma$ convolutions (kernel: 1, no padding). $g_{enc}^3$ has one convolution (kernel: 4, stride: 2, padding: 1) followed by $\mu$ and $\sigma$ convolutions (kernel: 1, no padding). The final module, $g_{ar}$, is a GRU with an output of size 64 $\times$ 256. ReLU is applied after each convolution except in the $\mu$ and $\sigma$ layers. The total downsampling factor is 160, producing a feature vector for every 10 ms of speech. Batch norm is applied to LibriSpeech but not to the artificial dataset. While it wasn't strictly necessary for LibriSpeech, it significantly increased training speed, which is beneficial given the dataset's large size. All modules are trained in parallel without gradients flowing between modules.

\subsubsection{Training Procedure} SIM is trained with the Adam optimizer (learning rate: $2 \times 10^{-4}$, batch size: 8). The maximum number of patches to predict in the future $K$ is set to 10, with 1000 epochs on the artificial dataset and 100 on LibriSpeech. The regularization weight $\beta$ is set to 0.01 on the artificial dataset to encourage interpretability and 0.001 on LibriSpeech to balance interpretability and performance. Implementation details regarding drawing negative samples for $\fkmblank$ remain identical to the audio experiment from~\citep{lowePuttingEndEndtoEnd2020a}.

\subsection{Classification Performance} \label{cha:classific_performance}
We evaluate SIM's representations by training a fully connected linear layer on top of SIM's frozen pretrained backbone for classification tasks. Classifiers are trained on temporally-average-pooled representations for 10 epochs using Cross-Entropy and the Adam optimizer ($lr=0.001$). Tasks include vowel (3 labels) and syllable (9 labels) classification on the artificial dataset, and phoneme (41 labels) and speaker (251 labels) classification on LibriSpeech.

\begin{table} 
    \caption{Accuracy for classification tasks on the artificial speech and LibriSpeech datasets. $^a$Baseline results from~\cite{lowePuttingEndEndtoEnd2020a}.
    } \label{tbl:combined_accuracies}
    \centering
    \scriptsize 
    \begin{tabular}{lcc}
        \toprule
        \multicolumn{3}{l}{\textbf{Artificial Speech Dataset}} \\
        Method              & Vowel Classification (\%)   & Syllable Classification (\%) \\
        \midrule
        Supervised          & 91.19 $\pm$ 1.56            & 83.32 $\pm$ 2.06             \\
        Random Backbone     & 32.44 $\pm$ 4.44            & 9.88 $\pm$ 2.12              \\
        GIM                 & 95.24 $\pm$ 0.60            & 50.0 $\pm$ 1.55              \\
        SIM                 & 92.58 $\pm$ 2.06            & 44.53 $\pm$ 1.77             \\
        \midrule
        \multicolumn{3}{l}{\textbf{LibriSpeech Dataset}} \\
        Method              & Speaker Classification (\%) & Phone Classification (\%)    \\
        \midrule
        Supervised$^a$      & 98.90                       & 77.70                        \\
        Random Backbone$^a$ & 1.90                        & 27.60                        \\
        GIM                 & 98.60                       & 61.80                        \\
        SIM                 & 96.02                       & 60.22                        \\
        \bottomrule
    \end{tabular}
\end{table}

\noindent \textit{Results.} A known drawback of $\beta$-VAE regularization is increased performance degradation, as greater emphasis is placed on disentanglement through the hyperparameter $\beta$~\cite{kim2018disentangling}. Interestingly, Table~\ref{tbl:combined_accuracies} shows that this trade-off is quite manageable, especially considering that SIM applies this regularizer across different layers. Both SIM and CPC achieve high accuracy on speaker (96.02\%, 98.60\%) and vowel (92.58\%, 95.24\%) classification but perform worse on syllable (44.53\%, 50.00\%) and phoneme (60.22\%, 61.80\%) classification. This suggests that the InfoNCE objective favors global sequence features while preserving less local information. Adding a hidden layer improved training accuracy for the syllable task but did not improve test performance, indicating that consonant information may no longer be fully retained in the representations. Meanwhile, the randomly initialized backbone performs poorly across all tasks, confirming that SIM learns meaningful representations.

\subsection{Representation Analysis}

\subsubsection{Qualitative Assessment of Latent Space Smoothness}
To gain a notion of the smoothness of SIM's latent space, we train a decoder $D(\zt^3) = \tildex_t$ on the artificial dataset, using representations from $g_{enc}^3(\cdot)$ to decode interpolations between two latent representations. Two audio signals, ``bidi'' and ``baga'' are encoded into their respective latent representations, $\zbidi$ and $\zbaga$ ($64 \times 512$). Interpolated representations ${\z^3_\alpha} = (1 - \alpha) \zbidi + \alpha {\z^3}_{\text{baga}}$ are decoded for values of $\alpha$ between 0 and 1.

\noindent \textit{Results.} Fig.~\ref{fig:interpolate_dims} shows the decoded signals smoothly transitioning as $\alpha$ varies, with no abrupt changes or nonsensical outputs. Exploring other interpolated audio signals than the one presented here, is possible via our demo. When decoding real samples (non-interpolated), we noted that vowel sounds were consistently correct, but consonants were often unclear or incorrect, which aligns with our discussion in~\ref{cha:classific_performance} that consonant information may be less represented.

\noindent 
\begin{figure*}
\centering
\input{graphs/bidi_to_baga_full_interpol/audio_partial_interpolation_9SIM_512_bididu_1_to_bagaga_1_reconstr_loss=DecoderLossMSE}\hspace{-5mm}
\input{graphs/bidi_to_baga_full_interpol/audio_partial_interpolation_6SIM_512_bididu_1_to_bagaga_1_reconstr_loss=DecoderLossMSE}\hspace{-5mm}
\input{graphs/bidi_to_baga_full_interpol/audio_partial_interpolation_3SIM_512_bididu_1_to_bagaga_1_reconstr_loss=DecoderLossMSE}\hspace{-5mm}
\input{graphs/bidi_to_baga_full_interpol/audio_partial_interpolation_0SIM_512_bididu_1_to_bagaga_1_reconstr_loss=DecoderLossMSE}
\caption{Interpolated latent representations between two audio signals (``bidi'' and ``baga'') using SIM. Each plot shows the decoded signal for different interpolation factors $\alpha$. Listen to the decoded audio, among others, on \href{https://colab.research.google.com/drive/1-4PxKfgBcEuPlSdNKE0fu24nV-bvIN2V}{\underline{Google Colab}}.} \label{fig:interpolate_dims} 
\end{figure*}


\subsubsection{Quantitative Evaluation of Specific Information Spread} \label{cha:experim_vowel_concentration}
To assess how vowel and speaker information is distributed across latent dimensions, we train linear classifiers (without bias) on average-pooled representations from $g_{enc}^1(\cdot)$, $g_{enc}^2(\cdot)$, and $g_{enc}^3(\cdot)$. Classifier weights indicate the relevance of each dimension, with large magnitudes signifying high importance.

\noindent \textit{Results.} As shown in Fig.~\ref{fig:distr weights classif}, SIM concentrates vowel/speaker information in fewer dimensions, which is beneficial for interpretability. GIM, on the other hand, spreads this information more broadly thereby requiring a larger number of neurons to be studied. Accuracy for vowel identification in GIM: 95.94\%, 92.81\%, 94.06\%, and in SIM: 87.5\%, 93.44\%, 91.87\%. For speaker identification, GIM: 88.29\%, 97.83\%, 98.56\%, and SIM: 68.36\%, 91.76\%, 94.32\%.
\begin{figure}
    \centering
    \input{graphs/density_plots/de_boer/Module_0_plot_of_512_dimensions_kde_plot_512_dims}
    \hspace{-5mm}%
    \input{graphs/density_plots/de_boer/Module_1_plot_of_512_dimensions_kde_plot_512_dims}
    \hspace{-5mm}%
    \input{graphs/density_plots/de_boer/module_2_plot_of_512_dimensions_kde_plot_512_dims}
    \hspace{-1em}%
    \input{graphs/density_plots/libri/Module_0_plot_of_512_dimensions_kde_plot_512_dims}
    \hspace{-5mm}%
    \input{graphs/density_plots/libri/Module_1_plot_of_512_dimensions_kde_plot_512_dims}
    \hspace{-5mm}%
    \input{graphs/density_plots/libri/module_2_plot_of_512_dimensions_kde_plot_512_dims} \\
    \caption{
        Distribution of linear classifier weights for vowel prediction (artificial dataset, left) and speaker identification (LibriSpeech, right), trained on representations from $g_{enc}^1(\cdot), g_{enc}^2(\cdot), g_{enc}^3(\cdot)$. SIM's classifiers show more weights near zero, indicating that vowel and speaker information is concentrated in fewer dimensions. }
    \label{fig:distr weights classif}
\end{figure}

\subsubsection{Quantitative Evaluation of General Information Spread}
To further observe the impact on interpretability through unit analysis and to analyze whether our representations align with the known disentanglement properties from $\beta$-VAE's regularizer~\cite{higginsBetaVAELearningBasic2022, chen2018isolating, kim2018disentangling}, we introduce a metric to measure how effectively a decoder $D:\mathcal{Z}^m \rightarrow \mathcal{X}$ can reconstruct a target signal when only the most relevant latent dimensions are modified. This serves as a proxy for entanglement, especially given that the artificial dataset has only a few ground truth factors, requiring far fewer dimensions than the available 512.

For each pair of starting and target representations $\left( \zmstart, \zmtarget \right)$, we define an interpolated representation, $\zmalpha$, where only the $N$ most important dimensions (those with the greatest average difference over 64 time frames) are altered to match $\zmtarget$. The similarity between the decoded $\zmalpha$ and $\zmtarget$ measures how many dimensions are needed to transform the signal. The relative error is computed as:

\begin{equation}
    \delta = \frac{ \text{MAE}\left( D(\zmtarget), D(\zmalpha) \right) }{\text{MAE}\left( D(\zmstart), D(\zmalpha) \right)}
\end{equation}
Here, $\delta$ ranges from 0 (exact match to target) to 1 (no effect from altering dimensions).

\noindent\textit{Decoder Details.} Each decoder $D(\zt^m)$ is trained for 50 epochs on representations from a specific module $g_{enc}^m(\cdot)$, using the MSE loss, a learning rate of $2 \times 10^{-4}$, and batch size 8. The decoder mirrors the encoder architecture, but for the first module, two additional layers (kernel size 3, padding 1, stride 1) were added to improve reconstruction.

\noindent\textit{Results.} Table~\ref{tbl:partial_interpol_errors} shows the relative errors. Across all depths, SIM reconstructs the target signal using fewer dimensions than GIM. For the artificial dataset, GIM needs at least half of the dimensions for successful reconstruction, whereas SIM achieves similar results with just 1/8th. Given the limited information in this dataset, which theoretically requires far fewer than 512 dimensions, multiple of GIM's dimensions seem sensitive to similar attributes. This implies a more entangled representation, which aligns with earlier findings in Fig.~\ref{fig:distr weights classif}. For LibriSpeech, SIM consistently requires fewer dimensions, averaging around half the number used by GIM, showing potential to scale well to more complex datasets.

\begin{table*}
    \caption{Relative reconstruction error $\delta$ (\%) when only the $N$ most important dimensions out of 512 are active. Lower values are preferred. GIM distributes relevant information across more dimensions than SIM.} \label{tbl:partial_interpol_errors}
    \centering
    \small
    \setlength{\tabcolsep}{3.5pt}
    \renewcommand{\arraystretch}{1.2}
   \resizebox{1.0\columnwidth}{!}{%
    \begin{tabular}{c|c|ccccccccc|ccccccccc}
        \toprule
        \multirow{2}{*}{\textbf{Module}} & \multirow{2}{*}{\textbf{Method}} & \multicolumn{9}{c|}{\textbf{Artificial Speech Dataset}} & \multicolumn{9}{c}{\textbf{LibriSpeech Dataset}} \\
        \cmidrule(lr){3-11} \cmidrule(lr){12-20}
        & & \textbf{2} & \textbf{4} & \textbf{8} & \textbf{16} & \textbf{32} & \textbf{64} & \textbf{128} & \textbf{256} & \textbf{512} & \textbf{2} & \textbf{4} & \textbf{8} & \textbf{16} & \textbf{32} & \textbf{64} & \textbf{128} & \textbf{256} & \textbf{512} \\
        \midrule
        
        \multirow{2}{*}{$g_{enc}^1$} 
        & GIM & \cellcolor[rgb]{0.83,0.50,0.57}99.32 & \cellcolor[rgb]{0.84,0.51,0.57}98.71 & \cellcolor[rgb]{0.85,0.52,0.57}97.6 & \cellcolor[rgb]{0.87,0.54,0.58}95.46 & \cellcolor[rgb]{0.91,0.58,0.58}91.2 & \cellcolor[rgb]{0.97,0.69,0.62}82.35 & \cellcolor[rgb]{1.00,0.91,0.75}62.96 & \cellcolor[rgb]{0.75,0.89,0.70}24.33 & \cellcolor[rgb]{0.50,0.70,0.61}0
        & \cellcolor[rgb]{0.87,0.54,0.58}95.55 & \cellcolor[rgb]{0.90,0.58,0.58}91.77 & \cellcolor[rgb]{0.94,0.62,0.59}87.47 & \cellcolor[rgb]{0.97,0.70,0.62}81.12 & \cellcolor[rgb]{0.99,0.82,0.68}71.65 & \cellcolor[rgb]{1.00,0.95,0.79}58.02 & \cellcolor[rgb]{0.92,0.97,0.77}39.52 & \cellcolor[rgb]{0.65,0.85,0.68}16.8 & \cellcolor[rgb]{0.50,0.70,0.61}0 \\
        
        & SIM & \cellcolor[rgb]{0.95,0.66,0.61}84.01 & \cellcolor[rgb]{1.00,0.91,0.75}62.5 & \cellcolor[rgb]{0.90,0.96,0.75}37.42 & \cellcolor[rgb]{0.82,0.92,0.71}29.45 & \cellcolor[rgb]{0.74,0.89,0.70}23.64 & \cellcolor[rgb]{0.66,0.85,0.69}17.96 & \cellcolor[rgb]{0.58,0.81,0.66}11.97 & \cellcolor[rgb]{0.53,0.76,0.63}5.66 & \cellcolor[rgb]{0.50,0.70,0.61}0
        & \cellcolor[rgb]{0.93,0.62,0.59}88.02 & \cellcolor[rgb]{0.98,0.71,0.63}79.83 & \cellcolor[rgb]{1.00,0.89,0.73}64.92 & \cellcolor[rgb]{0.89,0.96,0.75}36.93 & \cellcolor[rgb]{0.63,0.84,0.68}15.55 & \cellcolor[rgb]{0.56,0.80,0.66}10.62 & \cellcolor[rgb]{0.53,0.77,0.64}6.96 & \cellcolor[rgb]{0.52,0.74,0.63}3.53 & \cellcolor[rgb]{0.50,0.70,0.61}0 \\
        
        \midrule
        
        \multirow{2}{*}{$g_{enc}^2$} 
        & GIM & \cellcolor[rgb]{0.83,0.51,0.57}98.86 & \cellcolor[rgb]{0.84,0.52,0.57}97.91 & \cellcolor[rgb]{0.86,0.53,0.58}96.17 & \cellcolor[rgb]{0.89,0.56,0.58}93.07 & \cellcolor[rgb]{0.93,0.62,0.59}87.58 & \cellcolor[rgb]{0.98,0.74,0.64}77.8 & \cellcolor[rgb]{1.00,0.94,0.77}59.81 & \cellcolor[rgb]{0.79,0.91,0.70}27.46 & \cellcolor[rgb]{0.50,0.70,0.61}0
        & \cellcolor[rgb]{0.85,0.52,0.57}97.35 & \cellcolor[rgb]{0.87,0.55,0.58}94.88 & \cellcolor[rgb]{0.91,0.58,0.58}90.82 & \cellcolor[rgb]{0.95,0.66,0.61}84.7 & \cellcolor[rgb]{0.98,0.75,0.65}76.6 & \cellcolor[rgb]{1.00,0.87,0.71}67.02 & \cellcolor[rgb]{1.00,0.97,0.83}54.25 & \cellcolor[rgb]{0.84,0.93,0.72}31.32 & \cellcolor[rgb]{0.50,0.70,0.61}0 \\
        
        & SIM & \cellcolor[rgb]{0.92,0.59,0.58}89.89 & \cellcolor[rgb]{0.97,0.69,0.62}81.78 & \cellcolor[rgb]{1.00,0.88,0.72}65.96 & \cellcolor[rgb]{0.92,0.97,0.77}39.26 & \cellcolor[rgb]{0.80,0.91,0.70}28.04 & \cellcolor[rgb]{0.71,0.87,0.70}20.98 & \cellcolor[rgb]{0.61,0.83,0.67}14.0 & \cellcolor[rgb]{0.53,0.77,0.64}6.97 & \cellcolor[rgb]{0.50,0.70,0.61}0
        & \cellcolor[rgb]{0.93,0.61,0.58}88.75 & \cellcolor[rgb]{0.97,0.69,0.62}82.32 & \cellcolor[rgb]{0.99,0.78,0.66}74.74 & \cellcolor[rgb]{1.00,0.88,0.72}65.83 & \cellcolor[rgb]{1.00,1.00,0.87}49.4 & \cellcolor[rgb]{0.85,0.94,0.72}32.47 & \cellcolor[rgb]{0.73,0.88,0.70}22.44 & \cellcolor[rgb]{0.58,0.81,0.66}12.39 & \cellcolor[rgb]{0.50,0.70,0.61}0 \\
        
        \midrule
        
        \multirow{2}{*}{$g_{enc}^3$} 
        & GIM & \cellcolor[rgb]{0.83,0.51,0.57}99.01 & \cellcolor[rgb]{0.84,0.51,0.57}98.16 & \cellcolor[rgb]{0.85,0.53,0.58}96.59 & \cellcolor[rgb]{0.88,0.56,0.58}93.9 & \cellcolor[rgb]{0.92,0.60,0.58}89.29 & \cellcolor[rgb]{0.97,0.70,0.62}81.09 & \cellcolor[rgb]{1.00,0.89,0.73}65.23 & \cellcolor[rgb]{0.84,0.93,0.72}31.76 & \cellcolor[rgb]{0.50,0.70,0.61}0
        & \cellcolor[rgb]{0.88,0.55,0.58}94.37 & \cellcolor[rgb]{0.91,0.58,0.58}90.65 & \cellcolor[rgb]{0.95,0.65,0.60}85.08 & \cellcolor[rgb]{0.98,0.74,0.64}78.1 & \cellcolor[rgb]{0.99,0.82,0.68}71.47 & \cellcolor[rgb]{1.00,0.88,0.72}65.68 & \cellcolor[rgb]{1.00,0.96,0.80}57.0 & \cellcolor[rgb]{0.89,0.95,0.75}35.95 & \cellcolor[rgb]{0.50,0.70,0.61}0 \\
        
        & SIM & \cellcolor[rgb]{0.89,0.57,0.58}92.74 & \cellcolor[rgb]{0.94,0.62,0.59}87.44 & \cellcolor[rgb]{0.98,0.74,0.65}77.56 & \cellcolor[rgb]{1.00,0.94,0.78}59.27 & \cellcolor[rgb]{0.93,0.97,0.77}39.89 & \cellcolor[rgb]{0.80,0.91,0.70}27.96 & \cellcolor[rgb]{0.66,0.85,0.68}17.55 & \cellcolor[rgb]{0.54,0.77,0.64}7.73 & \cellcolor[rgb]{0.50,0.70,0.61}0
        & \cellcolor[rgb]{0.94,0.63,0.59}86.76 & \cellcolor[rgb]{0.97,0.70,0.62}81.14 & \cellcolor[rgb]{0.99,0.77,0.66}75.54 & \cellcolor[rgb]{0.99,0.83,0.68}70.96 & \cellcolor[rgb]{1.00,0.89,0.73}64.75 & \cellcolor[rgb]{1.00,0.99,0.85}52.02 & \cellcolor[rgb]{0.92,0.97,0.77}39.23 & \cellcolor[rgb]{0.74,0.89,0.70}23.38 & \cellcolor[rgb]{0.50,0.70,0.61}0 \\
        \bottomrule
    \end{tabular}
    }
\end{table*}

\subsubsection{Qualitative Analysis of Latent Shape}
To evaluate the structure of SIM's latent space, we first compute the representations $\z_t^3 = g_{enc}^3(\xt)$ for each sample from the test set. For each of their 512 dimensions, we construct a histogram with 100 bins, showing how all activations of the dataset for an individual dimension are spread. Figure~\ref{fig:distrib_activations_per_dim} displays the histograms for a selected set of dimensions. SIM's activations consistently follow a Gaussian distribution, aligning with our design goal of regularizing the latent space toward a standard normal distribution. This predictable structure helps post-hoc interpretability tools by clearly describing the regions of interest. In contrast, the latent representations produced by GIM show greater variation. In particular, dimension 9 for the artificial dataset, and dimensions 13 and 14 for LibriSpeech are shifted away from the origin or differ from the other dimensions.
Note that in GIM's current implementation, the latent spaces are implicitly constrained to center around the origin due to the use of a bias-free discriminator in Eq.~\ref{eq:fkm}. If the discriminator were non-linear or had a bias term, the shape of the latent space could potentially be even less predictable.

\begin{figure}
    \centering
    \input{graphs/latent_space_shape/gim_de_boer/distribution_latent_space_MEAN_SIM_dim=0}
    \hspace{-1.5em}%
    \input{graphs/latent_space_shape/gim_de_boer/distribution_latent_space_MEAN_SIM_dim=1}
    \hspace{-1.5em}%
    \input{graphs/latent_space_shape/gim_de_boer/distribution_latent_space_MEAN_SIM_dim=2}
    \hspace{-1.5em}%
    \input{graphs/latent_space_shape/gim_de_boer/distribution_latent_space_MEAN_SIM_dim=3}
    \hspace{-1.5em}%
    \input{graphs/latent_space_shape/gim_de_boer/distribution_latent_space_MEAN_SIM_dim=8} \\
    \vspace{-1.2em}
    \input{graphs/latent_space_shape/sim_de_boer/distribution_latent_space_MEAN_SIM_dim=0}
    \hspace{-1.5em}%
    \input{graphs/latent_space_shape/sim_de_boer/distribution_latent_space_MEAN_SIM_dim=1}
    \hspace{-1.5em}%
    \input{graphs/latent_space_shape/sim_de_boer/distribution_latent_space_MEAN_SIM_dim=2}
    \hspace{-1.5em}%
    \input{graphs/latent_space_shape/sim_de_boer/distribution_latent_space_MEAN_SIM_dim=3}
    \hspace{-1.5em}%
    \input{graphs/latent_space_shape/sim_de_boer/distribution_latent_space_MEAN_SIM_dim=8} \\
    %
    %
    \vspace{-0.2em}
    \input{graphs/latent_space_shape/gim_libri/distribution_latent_space_MEAN_SIM_dim=0}
    \hspace{-1.5em}%
    \input{graphs/latent_space_shape/gim_libri/distribution_latent_space_MEAN_SIM_dim=1}
    \hspace{-1.5em}%
    \input{graphs/latent_space_shape/gim_libri/distribution_latent_space_MEAN_SIM_dim=2}
    \hspace{-1.5em}%
    \input{graphs/latent_space_shape/gim_libri/distribution_latent_space_MEAN_SIM_dim=12}
    \hspace{-1.5em}%
    \input{graphs/latent_space_shape/gim_libri/distribution_latent_space_MEAN_SIM_dim=13} \\
    \vspace{-1.2em}
    \input{graphs/latent_space_shape/sim_libri/distribution_latent_space_MEAN_SIM_dim=0}
    \hspace{-1.5em}%
    \input{graphs/latent_space_shape/sim_libri/distribution_latent_space_MEAN_SIM_dim=1}
    \hspace{-1.5em}%
    \input{graphs/latent_space_shape/sim_libri/distribution_latent_space_MEAN_SIM_dim=2}
    \hspace{-1.5em}%
    \input{graphs/latent_space_shape/sim_libri/distribution_latent_space_MEAN_SIM_dim=12}
    \hspace{-1.5em}%
    \input{graphs/latent_space_shape/sim_libri/distribution_latent_space_MEAN_SIM_dim=13} \\
    \caption{Top rows: Artificial Dataset, bottom: LibriSpeech. Distribution of activations per dimension. SIM's activations have a consistent shape across dimensions, whereas GIM enforces no latent space constraints, resulting in greater variation in certain dimensions. Other dimensions are available \href{https://github.com/fdenoodt/Smooth-InfoMax/blob/main/apdx/main.pdf}{\underline{here}}. }
    \label{fig:distrib_activations_per_dim}
\end{figure}

    \section{Related Work}
This work uses the benefits of $\beta$-VAE's regularization, applying them across multiple layers to improve post-hoc analysis of the network.

In terms of \textit{interpretability through regularization}, various approaches have been explored. For instance, sparsity regularization in the activations of hidden representations~\cite{vincent2008extracting, bengio_learning_deep_architectures, xavier_deep_sparse_rectified_nns} has been shown to improve the compression of information into fewer dimensions, reducing the number of neurons to analyze, yet this does not encourage disentanglement or smoothness. Similarly,~\cite{moshe_improving_inerpret_regul_neural_activ} improves model interpretability through regularization of activations, modifying the model's behavior such that existing explainability methods produce explanations that better align with human perception. Another notable method involves tree regularization~\cite{wuOptimizingInterpretabilityDeep2021}, which constrains neural networks to be well-approximated by decision trees, thereby improving interpretability. While effective, its applications are typically limited to simpler tasks where decision trees can easily be formed. In contrast, SIM relies on post-hoc tools for its analysis but is able to model complex audio or vision tasks. Regarding disentanglement, traditional methods have typically focused on regularizing a single layer in the architecture~\cite{higginsBetaVAELearningBasic2022, chen2018isolating, kim2018disentangling, weining2017unsupervised, ge_encouraging_disentangled_convex_repr}, whereas SIM applies one of these approaches in a new context and across multiple layers.

For \textit{post-hoc interpretability}, SIM’s decoder is similar to approaches like~\cite{erhanVisualizingHigherLayerFeatures2009, simonyanDeepConvolutionalNetworks2014}, which reconstruct inputs from latent representations but rely on gradient ascent rather than a learned decoder. However, since NNs are not typically bijective, the reconstructions found do not necessarily map to a similar point from the dataset, resulting in often noisy and unclear reconstructions~\cite{erhanVisualizingHigherLayerFeatures2009, simonyanDeepConvolutionalNetworks2014}. To improve intelligibility, recent feature-activation-based methods incorporate human priors~\cite{mai_nguyen_synthes_preferr_inpus_neurons_deep_generator_networks, Mordvintsev_inceptionism, mahendran_visual_deep_conv_natural_preimgs}, guiding reconstructions toward more interpretable outputs. Decoders offer an alternative by directly learning to reconstruct inputs, implicitly encoding priors from the training data. However, they may introduce hallucinations that do not fully reflect the original model. Nonetheless, both gradient-based and decoder-based approaches could benefit from SIM’s structured latent spaces due to their encouraged smoothness, better disentanglement, and well-defined shapes.

    \section{Discussion}
We presented Smooth InfoMax, a self-supervised representation learning approach that incorporates interpretability requirements into the design of the model. Our proposal demonstrates how $\beta$-VAE regularization can be integrated into GIM's contrastive learning framework at various depths in the network. As such, SIM enjoys GIM's computational advantages—such as decoder-less training, large-scale distributed training for architectures that would otherwise not fit in memory, and reduced vanishing gradients—while also preserving the well-structured latent space properties of $\beta$-VAEs across layers. Remarkably, this is achieved without significantly compromising performance.
SIM enables more effective post-hoc interpretability, bringing us closer to understanding the internal workings of these neural networks.

\noindent
\textbf{Limitations and Future Work}
Although the latent space properties of $\beta$-VAEs improve post-hoc interpretability, the overall success still depends on the faithfulness of the generated explanations and the clarity of the information encoded in the representations. Additionally, SIM shows a small performance gap relative to its baseline, suggesting that integrating recent advances in disentanglement could be beneficial. While our evaluation focuses on sequential speech data (an XAI domain less exhaustively explored than vision), SIM’s InfoNCE-based architecture is easily adaptable to other modalities, such as vision and natural language.
Finally, SIM could be valuable beyond GIM as its probabilistic architecture and regularization can be integrated into other frameworks too, including end-to-end NNs as shown in this \href{https://github.com/fdenoodt/Smooth-InfoMax/blob/main/apdx/sim_end_to_end_vision/main.ipynb}{\underline{toy example}}.

\section*{Credits}
This research was funded by the Department of Computer Science at the University of Antwerp and by the Vrije Universiteit Brussel.

%

    \bibliographystyle{unsrt}
    \bibliography{refs}

\end{document}